\documentclass{article}
\usepackage{xcolor}
\definecolor{mygrey}{HTML}{949494}
\usepackage{arxiv}
\usepackage{caption}
\DeclareCaptionLabelSeparator{pipe}{\enspace$\mid$\enspace} 
\usepackage[utf8]{inputenc} 
\usepackage[T1]{fontenc}    
\usepackage{url}            
\usepackage{booktabs}       
\usepackage{amsfonts}       
\usepackage{nicefrac}       
\usepackage{microtype}      
\usepackage{lipsum}
\usepackage{graphicx}
\usepackage{amsmath, amssymb}
\usepackage{hyperref}
\usepackage{cleveref}
\Crefname{figure}{Fig.}{Figs.}
\usepackage{graphicx}
\usepackage{orcidlink}
\usepackage{comment}
\usepackage{tcolorbox} 
\tcbuselibrary{theorems}
\usepackage{nameref}
\usepackage{enumitem}
\usepackage[numbers]{natbib}
\usepackage{array}
\newcolumntype{C}[1]{>{\centering\arraybackslash}p{#1}}
\newcolumntype{L}[1]{>{\raggedright\arraybackslash}p{#1}}

\definecolor{VUB_blauw}{rgb}{0.1529, 0.2667, 0.5529}
\usepackage{hyperref}       
\hypersetup{
    colorlinks,%
    citecolor=VUB_blauw,%
    filecolor=VUB_blauw,%
    linkcolor=VUB_blauw,%
    urlcolor=VUB_blauw
}
\graphicspath{ {./images/} }
\newcommand{\customCor}[1]{%
  \includegraphics[height=1em]{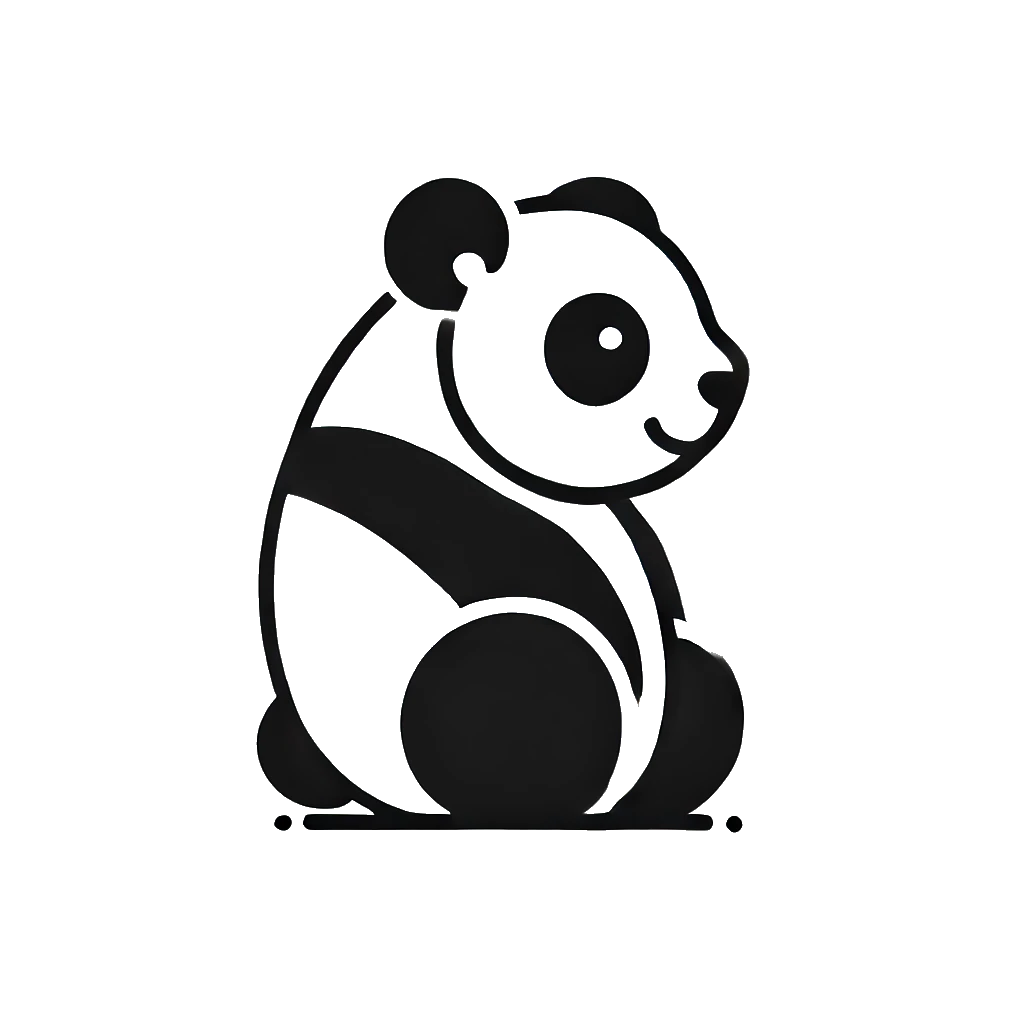} #1%
}

\AddToHook{shipout/background}{%
  \ifnum\value{page}=1 
    \pagenumbering{Roman} 
    \setcounter{page}{1} 
  \fi
  \ifnum\value{page}=2 
  \pagenumbering{arabic} 
  \setcounter{page}{2} 
  \fi
}

\title{The Relationship Between Reasoning and Performance in Large Language Models---o3 (mini) Thinks Harder, Not Longer.}
\runningtitle{o3 (mini) thinks harder, not longer}

\author{
  Marthe Ballon\textsuperscript{1,\customCor{ }} \\ 
  \orcidlinkc{0009-0000-4586-234X} \\
  \And
  Andres Algaba\textsuperscript{1} \\ 
  \orcidlinkc{0000-0002-0532-3066} \\
  \And
  Vincent Ginis\textsuperscript{1,2} \\ 
  \orcidlinkc{0000-0003-0063-9608} \\
  \and
  \textsuperscript{1}Data Analytics Lab, Vrije Universiteit Brussel, 1050 Brussel, Belgium \\
  \textsuperscript{2}School of Engineering and Applied Sciences, Harvard University, Cambridge, Massachusetts 02138, USA
}
    
\begin{document}
\maketitle
\renewcommand{\thefootnote}{} 
\footnotetext{\includegraphics[height=1em]{panda2.png} Corresponding author: marthe.ballon@vub.be}
\renewcommand{\thefootnote}{\arabic{footnote}} 
\thispagestyle{plain} 

\begin{abstract}
Large language models have demonstrated remarkable progress in mathematical reasoning, leveraging chain-of-thought and test-time compute scaling. However, many open questions remain regarding the interplay between reasoning token usage and accuracy gains. In particular, when comparing models across generations, it is unclear whether improved performance results from longer reasoning chains or more efficient reasoning. We systematically analyze chain-of-thought length across o1-mini and o3-mini variants on the Omni-MATH benchmark, finding that o3-mini (m) achieves superior accuracy without requiring longer reasoning chains than o1-mini. Moreover, we show that accuracy generally declines as reasoning chains grow across all models and compute settings, even when controlling for difficulty of the questions. This accuracy drop is significantly smaller in more proficient models, suggesting that new generations of reasoning models use test-time compute more effectively. Finally, we highlight that while o3-mini (h) achieves a marginal accuracy gain over o3-mini (m), it does so by allocating substantially more reasoning tokens across all problems, even the ones that o3-mini (m) can already solve. These findings provide new insights into the relationship between model capability and reasoning length, with implications for efficiency, scaling, and evaluation methodologies.
\end{abstract}

\keywords{chain-of-thought \and large language models \and reasoning models \and test-time compute \and mathematical benchmarks}

\section*{Introduction}
Large language models (LLMs) have evolved from handling basic natural language processing tasks to solving complex problems \citep{brown2020language,bubeck2023sparks,romera2024mathematical,trinh2024solving}. Scaling model size, data, and compute \citep{kaplan2020scaling} has enabled larger models to develop richer internal representations \citep{gurnee2024languagemodelsrepresentspace,hao2023reasoning} and emergent capabilities \citep{wei2022emergent}. Recently, a new class of reasoning models has emerged that couples reinforcement learning with test-time compute scaling \citep{muennighoff2025s1simpletesttimescaling,snell2024scaling}. These models leverage reasoning tokens to guide the chain-of-thought process and maintain coherence throughout complex problem-solving tasks~\citep{anderson2025phd,chen2024not,wang2025thoughts}. By explicitly optimizing the chain-of-thought in the reasoning tokens during training \citep{wei2022chain} and iteratively refining outputs at inference, these models achieve superior performance, including on challenging mathematical benchmarks \citep{guo2025deepseek,guan2025rstar}. Moreover, new test-time scaling laws demonstrate that longer reasoning—i.e. more reasoning tokens—yields log‑linear performance gains \citep{muennighoff2025s1simpletesttimescaling}.

In this paper, we examine whether more capable models within a single family (o-series of OpenAI) require a longer chain-of-thought to achieve higher performance or if they can reason more effectively. By systematically comparing the number of tokens in the chain-of-thought generated by o1‑mini, o3‑mini (m), and o3‑mini (h) on the Omni-MATH dataset \citep{gao2024omni}, we find that more proficient models (o1-mini vs. o3-mini (m)) do not generate longer reasoning chains to achieve higher accuracy. For all models and compute settings, we find that accuracy generally decreases as the chain-of-thought grows, even when controlling for question difficulty. This effect is notably smaller for more proficient models, indicating that o3-mini (m) tends to overthink less and uses reasoning tokens more effectively than o1-mini. However, within one model (o3-mini (m) vs. o3-mini (h)), we observe that the slower accuracy decrease per token is partially due to a higher average accuracy, but mainly due to the model allocating (more than) double the reasoning tokens for all questions.
Our findings contribute to the ongoing discussion about whether models such as o1 tend to overthink or underthink \citep{chen2024not,wang2025thoughts}, while complementing studies on reasoning step length \citep{jin2024impact}, input length \citep{levy2024same}, reasoning failure modes \citep{anderson2025phd}, and the optimization of mathematical reasoning \citep{zhong2024achieving}.

\begin{figure}[t]
	\centering
	\includegraphics{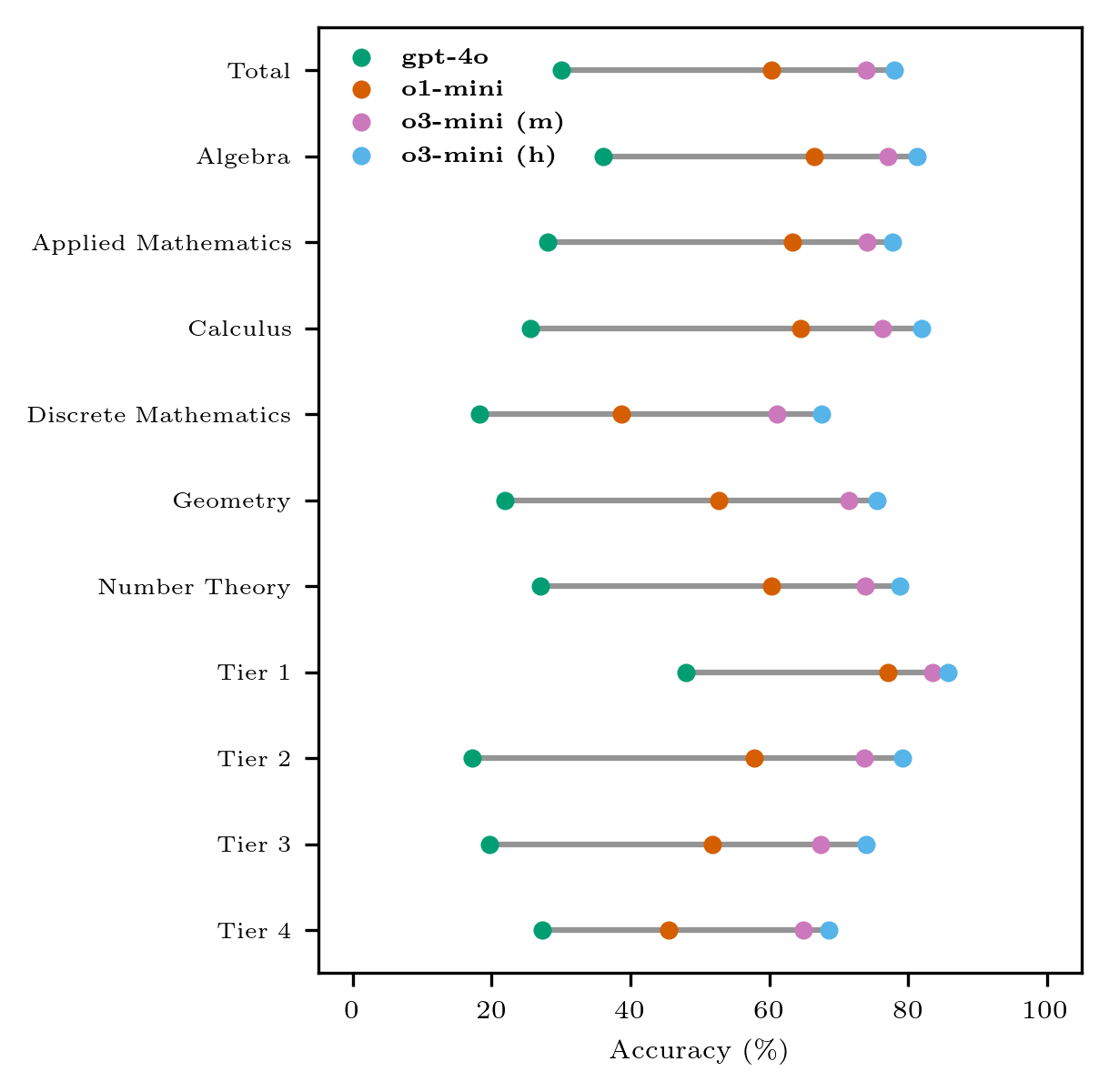}
	\caption{\textbf{Accuracy comparison of OpenAI models gpt-4o, o1-mini, o3-mini (m) and o3-mini (h) on the Omni-MATH benchmark.} This figure displays the accuracy of gpt-4o, o1-mini, o3-mini (m) and o3-mini (h) on the Omni-MATH benchmark across disciplines and difficulty tiers. The gpt-4o model fails to attain $50\%$ in any category and consistently lags behind the reasoning models. o1-mini significantly improves accuracy, reaching accuracies of $40$-$60\%$ across all domains, while the o3-models surpass $50\%$ accuracy in all categories. In general, accuracy declines as difficulty increases, with the exception of gpt-4o, which shows accuracy vs. difficulty level imbalance for Tiers $2$, $3$, and $4$.}
	\label{fig:main_1}
\end{figure}

We use the Omni-MATH dataset \citep{gao2024omni} to benchmark the reasoning abilities of o1-mini, o3-mini~(m) and o3-mini~(h). Existing benchmarks such as GSM8K \citep{cobbe2021training} and MATH \citep{hendrycks2021measuring} have become less effective in differentiating the mathematical abilities of LLMs due to the high accuracy rates they achieve on these tests. To address this, the Omni-MATH benchmark provides a rigorous evaluation framework at the Olympiad level. Unlike other benchmarks such as FrontierMath~\citep{glazer2024frontiermath}, GSM-symbolic \citep{mirzadeh2024gsm}, and sections of Humanity's Last Exam \citep{phan2025humanity}, the Omni-MATH dataset contains problems categorized into over 33 sub-domains and spanning more than 10 distinct difficulty levels. This detailed organization enables a nuanced assessment of LLMs' mathematical reasoning across various disciplines and complexities. In addition, the availability of Omni‑Judge \citep{gao2024omni}—a math‑evaluation model designed to verify and correct model‑generated answers against reference answers—ensures a straightforward and automated evaluation method. 

More general benchmarks like those in \citep{srivastava2022beyond} and MMLU \citep{hendrycks2020measuring}, along with specialized tests such as AI2 Reasoning \citep{clark2018think} and GPQA \citep{rein2023gpqa}, broaden the evaluation landscape to diverse reasoning domains. Additionally, coding benchmarks \citep{chen2021evaluating,jimenez2024swebench} highlight the importance of clearly defined reward models and verification systems in assessing models' performance. Future evaluations may consider additional benchmarks that incorporate broader language understanding or real-world reasoning challenges, though the current focus remains on math and coding due to the relative ease of implementing objective reward models and automated verification procedures in these domains.

\begin{figure}
	\centering
	\includegraphics{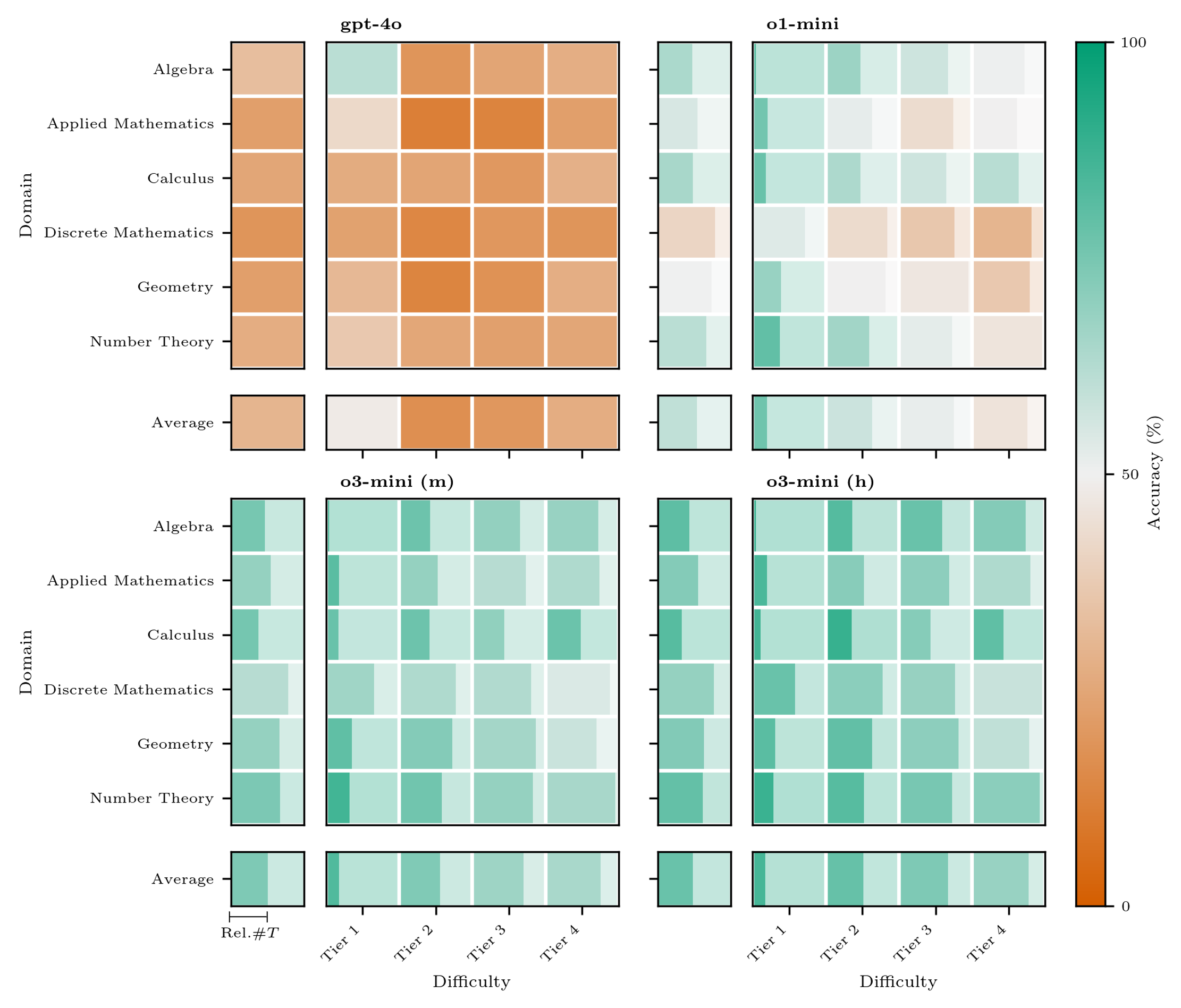}
	\caption{\textbf{Granular performance and reasoning token usage evaluation across domains and difficulty tiers of gpt-4o, o1-mini, o3-mini (m) and o3-mini (h) on the Omni-MATH benchmark.} The heatmaps visualize cross-sectional performance scores on a $0$-$100\%$ scale, represented by the color of the progress bar. The length of the progress bar in each cell represents relative token usage for the test-time scaled models. The extra column is computed by averaging over the rows. The extra row and ``average'' cell are computed independently to give equal weight to multi-domain questions (see \nameref{sec:Methods}). This figure shows that models allocate more computational resources to problems that require complex combinatorial reasoning (Geometry, Discrete Mathematics and Number Theory), whereas foundational arithmetic and algebra problems demand relatively fewer resources. On average, token usage scales with difficulty level.}
	\label{fig:main_2}
\end{figure}

\section*{Results}
Our data consists of $4428$ Olympiad-level math problems, the Omni-MATH benchmark, together with a reference answer and relevant metadata fields Domain and Difficulty (Appendix \Cref{fig:sample-1,fig:sample-2}). We consider six elementary mathematics domains, Algebra, Applied Mathematics, Calculus, Discrete Mathematics, Geometry and Number Theory (Appendix \Cref{fig:methodology-1}) and divide the data into four difficulty tiers, Tier $1$, Tier $2$, Tier $3$ and Tier $4$ (Appendix \Cref{fig:methodology-2}). Subsequently, we feed the problems to four OpenAI models, namely gpt-4o, o1-mini, o3-mini (m) and o3-mini (h) and make automated requests to the Omni-Judge model \cite{gao2024omni} to correct their answers (consult \nameref{sec:Methods} for more details).

\subsection*{Accuracy and relative token usage across models, disciplines, and difficulties}
\Cref{fig:main_1,fig:main_2} show the accuracy of OpenAI models gpt-4o, o1-mini, o3-mini (m) and o3-mini (h) across disciplines and difficulty tiers. The gpt-4o model performs consistently between $20\%$ and $30\%$ for all math disciplines but clearly lags behind the three reasoning models. o1-mini significantly improves accuracy in all categories, reaching $40$–$60\%$ on all domains. The introduction of o3-mini (m) further enhances performance, achieving $50\%$ in all categories. The o3-mini (h) model improves with approximately $4\%$ on average compared to o3-mini (m) and surpasses $80\%$ accuracy for Algebra and Calculus. A notable outlier is Discrete Mathematics, where performance deviates from the overall trend for all models. In general, accuracy declines as tier level increases. An exception is observed in gpt-4o, which performs better on Tier 4 than on Tiers 2 and 3 (the reasoning models also perform slightly better on Calculus Tier $3$ than Tier $4$). This anomaly suggests that the model might leverage unexpected heuristics or struggle disproportionately with mid-tier complexity.

Besides indicating accuracy (via the colors of the progress bars), \Cref{fig:main_2} also shows relative use of reasoning tokens (via the length of the progress bars) across the Omni-MATH dataset for o1-mini, o3-mini (m), and o3-mini (h). The relative use of tokens increases with the level of difficulty for all models, highlighting the need for computational resources for more difficult tasks. Discrete Mathematics stands out as a token-intensive domain, indicating a heavier combinatorial or multi-step reasoning load. Foundational mathematics areas such as Calculus and Algebra tend to consume fewer tokens, possibly because they are more procedurally straightforward. Interestingly, we observe that a relatively longer chain of reasoning does not generally lead to better performance, as many Tier $4$ math problems from token-intensive domains remain unsolved. Notable exceptions are the Geometry Tier $3$ problems, where all three reasoning models allocate more reasoning compute to Tier $3$ than Tier $4$ problems, resulting in a higher accuracy for Geometry Tier $3$ than Tier $4$.

\subsection*{Reasoning token distribution and performance vs. token usage}
\begin{figure}
	\centering
	\includegraphics{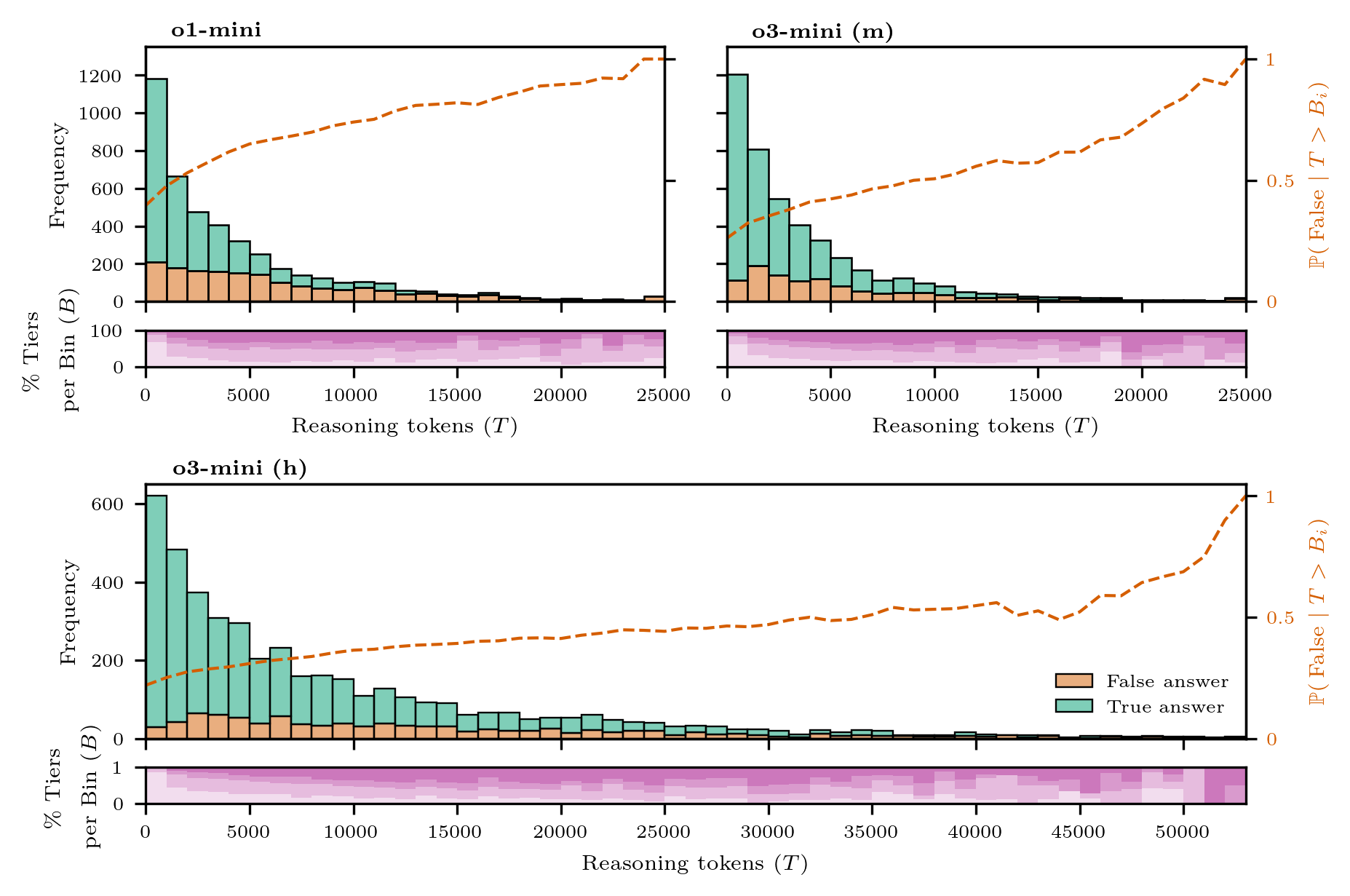}
	\caption{\textbf{Analysis of the reasoning token distribution, evolution of token region accuracy, and consistency between difficulty tiers and token usage for o1-mini, o3-mini (m) and o3-mini (h).} The main panels of the figure display the distribution of the reasoning tokens as a stacked histogram, illustrating the proportion of correctly and incorrectly answered questions in the Omni-MATH dataset by o1-mini, o3-mini (m) and o3-mini (h). The secondary $y$-axis depicts the probability that the model answers incorrectly given that the token count has surpassed the bin threshold (see \nameref{sec:Methods}). The panels below the histogram contain a filled histogram where the color opacity represents the difficulty level of the math questions (cfr. \Cref{fig:methodology-2}).
    The figure shows that o1-mini and o3-mini (m) have a similar reasoning token distribution, with o3-mini (m) giving more correct answers for high-token regions. o3-mini (h) has a good ratio of correct vs. incorrect answers, even for very high token counts. The probability of giving an incorrect answer increases with token count for all models. Finally, the relative proportion of tier levels in each bin reveal a clear transition from a region where the majority of the questions come from the lowest tiers to a region where the majority of the questions come from the highest tiers (for bins with a sufficient amount of data points). } 
	\label{fig:main_3}
\end{figure}

\begin{figure}
    \centering
    \includegraphics{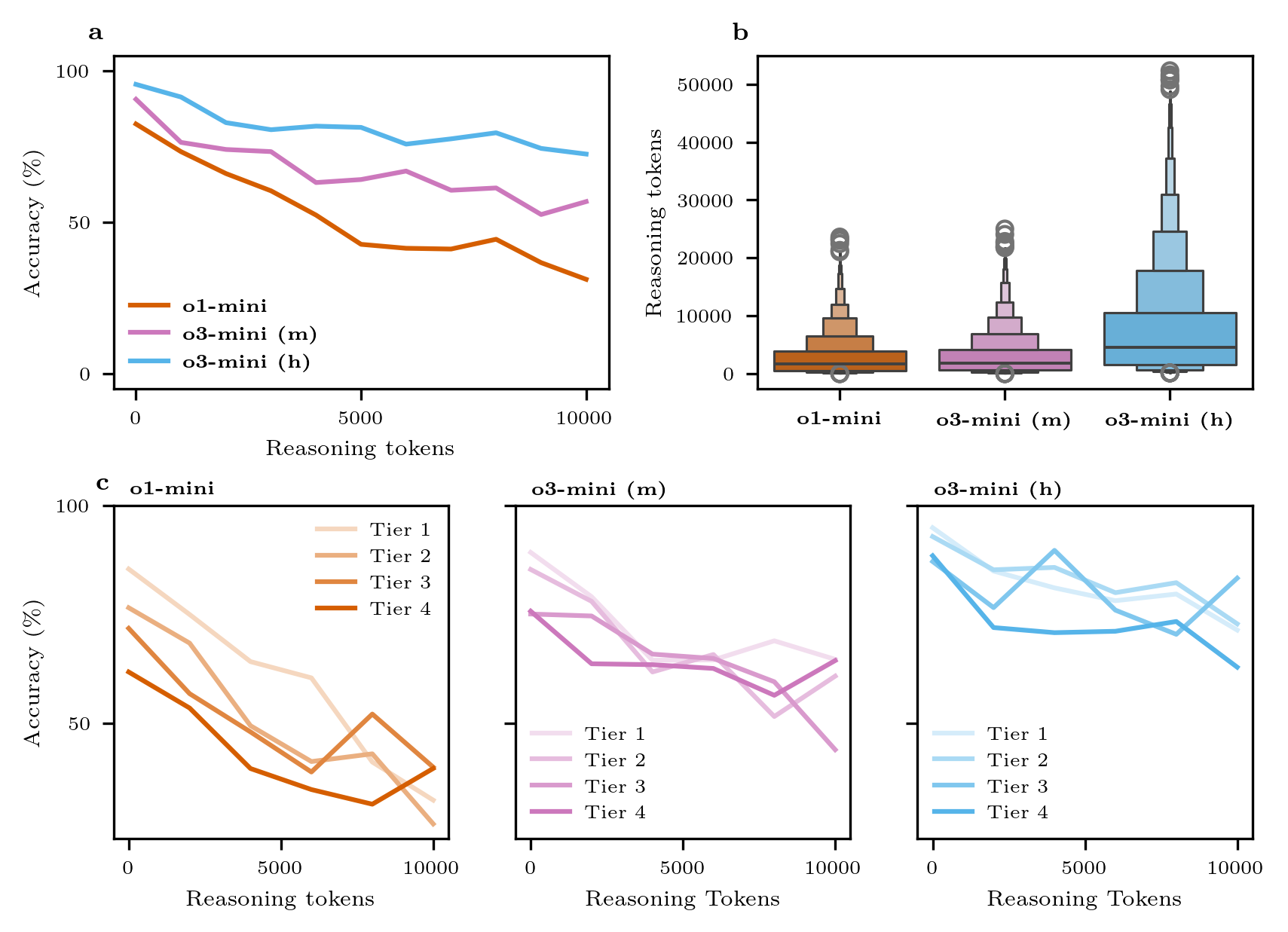}
    \caption{\textbf{o3 (mini) thinks harder, not longer.} This figure shows that o3-mini (m) does not require longer reasoning chains than o1-mini to achieve better accuracy and that, in general, more proficient models exhibit less accuracy decay as reasoning tokens increase. \textbf{a,} Accuracy per reasoning token, computed by dividing the number of correctly answered questions by the total number of questions in each bin of the histograms in \Cref{fig:main_3}. Accuracy declines as reasoning token usage increases. Furthermore, we observe that the slope of the lines becomes flatter for higher performing models. These effects are further quantified in the regression analysis (see \nameref{sec:Methods}). \textbf{b,} The boxplots show the distribution of the reasoning tokens for \textit{correctly} answered questions. Further investigation in the left panel of \Cref{fig:appendix-3} confirms that o1-mini and o3-mini (m) have a very similar token distribution. The token distribution of o3-mini (h) is stretched linearly with respect to the one of o3-mini (m) (\Cref{fig:appendix-3} right). 
    \textbf{c,} Stratifying plot \textbf{a} by difficulty level shows that, within difficulty tiers, accuracy also decreases with higher reasoning token usage. This suggests that the number of reasoning tokens, rather that difficulty level alone, can be used as a signal for the correctness of the model's answer.
    In \Cref{fig:appendix-2}, we show this also holds when stratifying across domains.
    }
    \label{fig:main_4}
\end{figure}

\Cref{fig:main_3,fig:main_4} display the relationship between the number of reasoning tokens and the performance of o1-mini, o3-mini (m), and o3-mini (h) on the Omni-MATH dataset (consult Appendix \Cref{fig:appendix-1} for gpt-4o analysis with completion tokens, which encompass both the tokens leading up to the answer and the answer itself). 
\Cref{fig:main_3} shows the proportion of the correct (green bars) versus incorrect (red bars) model responses across the reasoning token distribution. The red dashed line depicts the conditional error rate, i.e. the probability that the model answers incorrectly given the amount of used reasoning tokens (see \nameref{sec:Methods}). One first thing to note is that higher performing models have a better ratio of correct to incorrect answers, even for high token counts. This pattern is also reflected in the conditional error rate (red dashed line): the conditional error rate is almost instantly at $50\%$ for o1-mini whereas it takes about $12,000$ tokens for o3-mini (m) and $30,000$ for o3-mini (h) to reach a $50\%$ error rate. A second thing to note is that the token distributions of o1-mini and o3-mini (m) are very similar. \Cref{fig:main_4}b together with the left QQ-plot in \Cref{fig:appendix-3} further investigate this behavior by comparing the distribution of the reasoning tokens only for the questions that the models answered \emph{correctly}. Indeed, the almost identical token distributions show that o3-mini (m) does not use more reasoning tokens to achieve its superior performance to o1-mini on Omni-MATH. This suggests that o3-mini (m) reasons more effectively.
The token distribution of o3-mini (h) spans a significantly wider range of values, with the model allocating over $50,000$ reasoning tokens for some math problems. In addition, the right QQ-plot in \Cref{fig:appendix-3} shows that o3-mini (h) uses more reasoning tokens to solve all (correctly answered) questions, indicating that the small accuracy gain of $4\%$ compared to o3-mini (m) is accompanied by a large extra computational cost. 

 The panels below the histograms in \Cref{fig:main_3} display the relative proportion of tier levels in each bin. They reveal a clear transition from a region where the majority of the questions come from the lowest tiers to a region where the majority of the questions come from the highest tiers. Note that this pattern is visualized by the purple filled histograms in \Cref{fig:main_3} (higher token regions sometimes have insufficient data counts to show this pattern). The gradient confirms that more complex questions systematically demand greater reasoning depth, which is in line with prior observations. 
\Cref{fig:main_4}a shows that the average accuracy decreases with increasing use of reasoning tokens for all three models, but that this trend is the most pronounced for o1-mini and smaller for o3-mini (m) and o3-mini (h). While this could be attributed to higher-tier questions requiring more tokens, \Cref{fig:main_4}c shows that the trend remains even when stratifying by tier level. In \Cref{fig:appendix-2}, we show this also holds when stratifying across domains. This suggests that increased token usage, rather than question complexity alone, is related to accuracy. We use a logistic regression to quantify the effect size of using additional reasoning tokens on the probability of answering a question correctly, controlling for different levels of difficulty and domains (see \nameref{sec:Methods}). We report the average marginal effects in \Cref{tab:regression}. The accuracy decrease per 1000 reasoning tokens is $3.16\%$ for o1-mini, $1.96\%$ for o3-mini (m), and $0.81\%$ for o3-mini (h). These results indicate that while deeper reasoning is necessary for solving complex problems, there is a diminishing return, where excessive token usage correlates with reduced accuracy. However, this effect is smaller for higher performing models. Together with the results from the previous paragraph, we conclude that o3-mini (m) tends to overthink less and uses reasoning tokens more effectively than o1-mini. Within one model (o3-mini (m) vs. o3-mini (h)), the slower accuracy decrease per token is attributed to a stretched out token distribution along the $x$-axis and a shrunken distribution with respect to the $y$-axis.

\section*{Discussion}
By systematically comparing the number of tokens in the chain-of-thought generated by o1‑mini, o3‑mini (m), and o3‑mini (h) on the Omni-MATH dataset \citep{gao2024omni}, we find two important results. First, more proficient models (o1-mini vs. o3-mini (m)) do not require longer reasoning to achieve higher accuracy. Second, while accuracy generally declines with a longer chain-of-thought, this effect is notably smaller in more proficient models, underscoring that ``thinking harder'' is not the same as ``thinking longer''. A possible hypothesis for this accuracy drop is that models tend to reason more on problems they cannot solve. Another possibility is that longer reasoning chains inherently have a higher probability of leading to a wrong final solution, highlighting the need for mathematical benchmarks with reference reasoning templates. A practical takeaway from our study is that constraining the chain-of-thought (by setting \texttt{max\_completion\_tokens}) is more useful for weaker reasoning models than for stronger ones, as the latter still give a significant amount of correct answers for high-token regions.

The token count for o3-mini (h) contained the following subtlety: although o3-mini (h) solves additional problems compared to o3-mini (m), the model uses more tokens for all math problems. The slower decrease in accuracy per token is thus due to a stretched token distribution along the $x$-axis rather than a more effective usage of reasoning tokens. 

Our study relies on automated correction by Omni-Judge \citep{gao2024omni}, a model-based evaluator whose judgements could diverge from human corrections \citep{verga2024replacing,li2025preference}. Omni-Judge has only been validated for data leakage checks on o1-mini \citep{gao2024omni}; extending these checks to o3-mini remains future work, though we assume minimal overlap. Additionally, our prompting strategy employed here \citep{kojima2022zeroshot,wang2022self,yao2024tree} may not generalize to alternative approaches or more constrained prompt settings \citep{tam2024let}, and their interaction with test-time compute warrants further investigation. Many studies on prompting and reasoning were conducted on LLMs without test-time compute, so the broader implications for the latest generation of reasoning models \citep{guo2025deepseek} remain to be fully understood.

Our findings contribute to the debate on whether models such as o1 overthink or underthink \citep{chen2024not,wang2025thoughts}, and extend ongoing inquiries into reasoning step length \citep{jin2024impact}, input length \citep{levy2024same}, failure modes \citep{anderson2025phd}, and the optimization of mathematical reasoning \citep{zhong2024achieving}. By revealing how stronger models can achieve higher accuracy without proportionally longer chain-of-thought, we offer new insights into efficiency, scaling, and evaluation strategies. We also highlight the potential for next-generation test-time compute models to refine the balance between reasoning depth and resource usage.

\section*{Data and code availability}
Data associated with this study are publicly available in a public repository at \url{https://doi.org/10.5281/zenodo.14878936}. \\ The original Omni-MATH dataset is available at \url{https://huggingface.co/datasets/KbsdJames/Omni-MATH}. \\ The original Omni-Judge model is available at \url{https://huggingface.co/KbsdJames/Omni-Judge}. \\

The code for this publication is publicly available at \url{https://github.com/MartheBallon/analysis_o3-mini_thinks_harder_not_longer}. It is based on the Omni-MATH benchmark analysis code, publicly available at \url{https://github.com/KbsdJames/Omni-MATH}. We used Python 3.12.6 (\textit{pandas 2.2.3}, \textit{numpy 2.1.1}, \textit{matplotlib 3.9.2}, \textit{seaborn 0.13.2}, \textit{statsmodels 0.14.4}, and \textit{scikit-learn 1.5.2}) to analyse and visualize data and to conduct statistical analyses.

\section*{Acknowledgements}
This research received funding from the Flemish Government (AI Research Program).\\
Andres Algaba acknowledges a fellowship from the Research Foundation Flanders under Grant No.1286924N. \\ Vincent Ginis acknowledges support from Research Foundation Flanders under Grant No.G032822N and G0K9322N.

\section*{Author contributions}
Vincent Ginis and Marthe Ballon were responsible for the main idea of the study and designed the setup of the analysis. Andres Algaba was responsible for conducting the regression analysis. Marthe Ballon executed the analysis and designed the figures.  All authors discussed the results, and collaboratively drafted and revised the manuscript. 

\clearpage
\bibliographystyle{unsrt}

\setcounter{figure}{0}
\renewcommand{\thefigure}{A\arabic{figure}}
\setcounter{table}{0}
\renewcommand{\thetable}{A\arabic{table}}

\clearpage
\appendix

\renewcommand{\thesubsection}{\Alph{subsection}}

\section{Methods}
\label{sec:Methods}
We describe our experimental setup and provide the data processing details necessary to replicate our analysis. At the end of this section, we elaborate on the regression analysis conducted to analyse the effect size of increased reasoning token usage on accuracy.

\begin{figure}[t]
    \centering
    \includegraphics{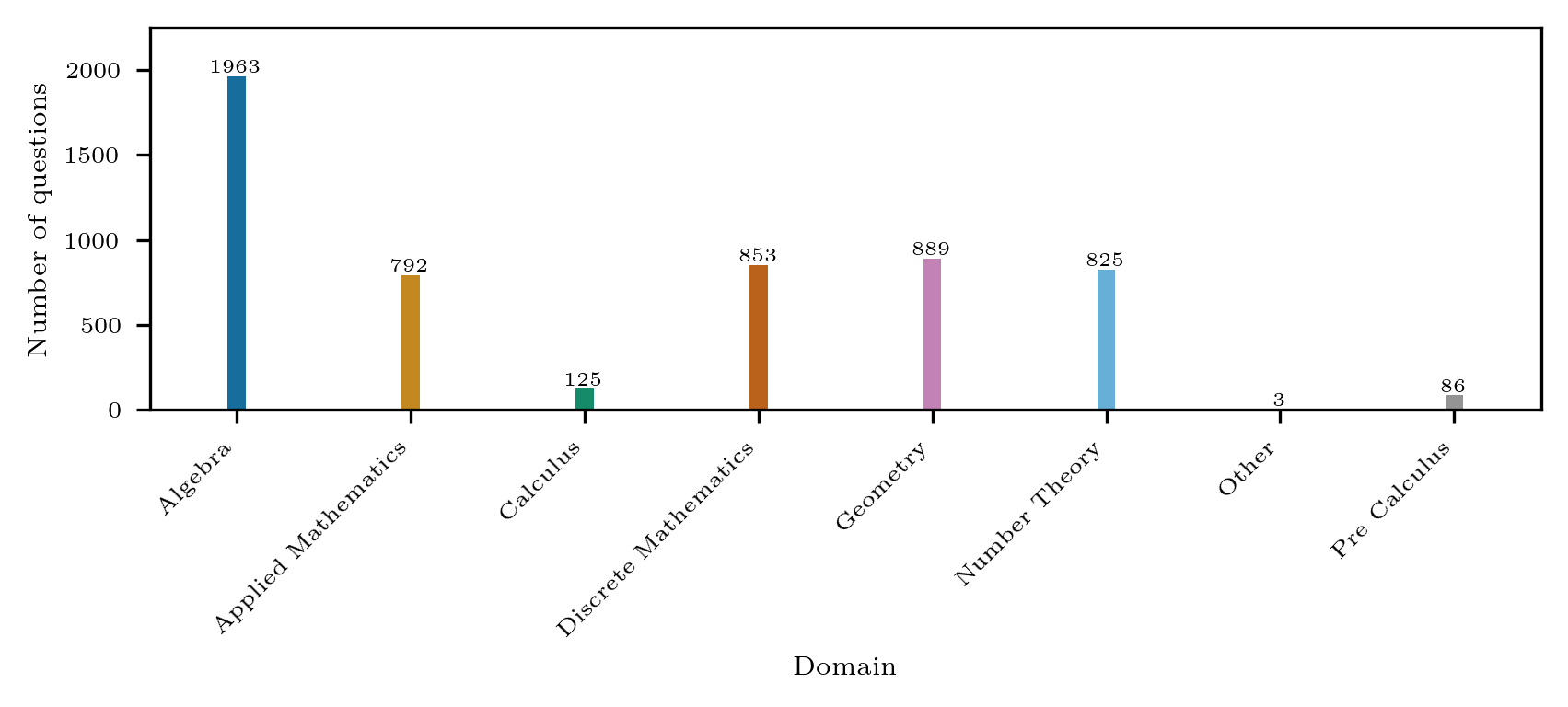}
    \caption{\textbf{Domain distribution of the Omni-MATH dataset.} This figure displays the distribution of the primary domains of the Omni-MATH dataset. Math problems that belong to multiple domains are counted for each domain, so the total number of question is higher than in \Cref{fig:methodology-2}.}.
    \label{fig:methodology-1}
\end{figure}

\begin{figure}[t]
    \centering
    \includegraphics{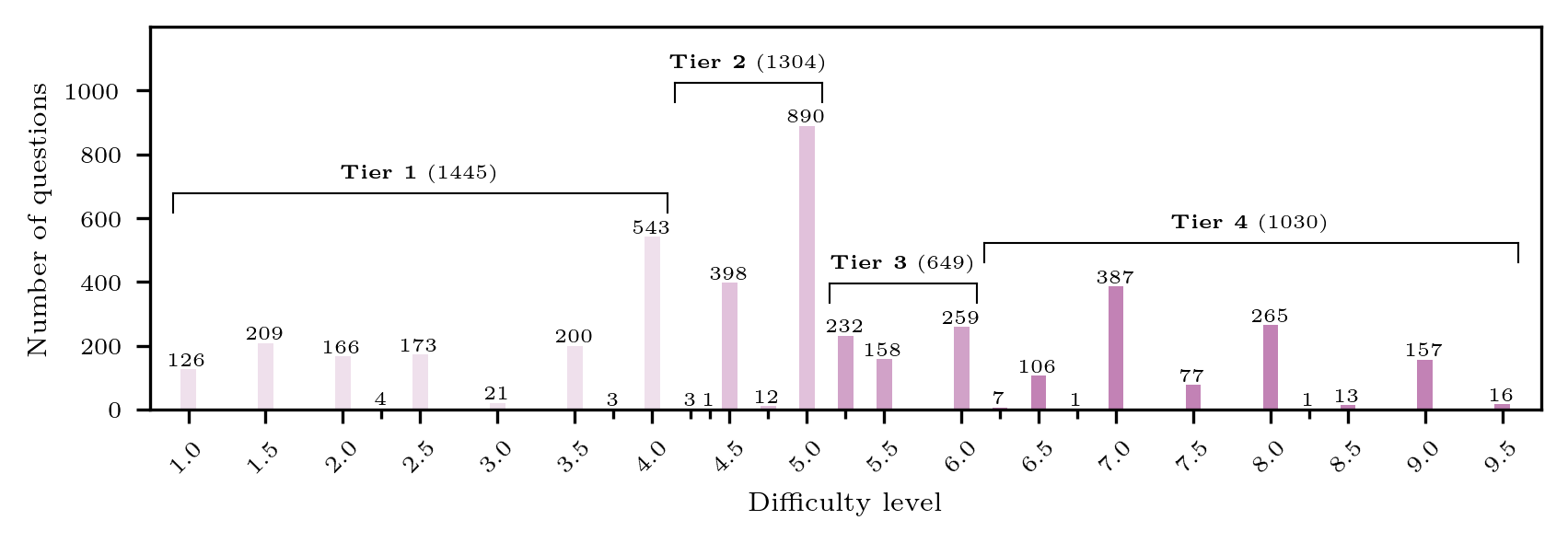}
    \caption{ \textbf{Classification of difficulty levels in balanced difficulty tiers.} This figure shows the difficulty distribution of the Omni-MATH dataset. The difficulty levels are classified in difficulty tiers based on the quartiles of the distribution (without separating difficulty levels). }
    \label{fig:methodology-2}
\end{figure}

\textbf{The Omni-MATH dataset}\quad The Omni-MATH benchmark \cite{gao2024omni} contains Olympiad-level math problems specifically designed to test the reasoning abilities of Large language models. Each entry in the dataset consists of a problem, an exact answer, and a written out solution together with the following metadata fields: Domain, Difficulty, and Source (see \Cref{fig:sample-1}). Each problem has between one and three domains of the form Mathematics $\rightarrow$ Primary domain $\rightarrow$ $\dots$, with a maximum length of five. In this paper, we only take the primary domains into account, as a more granular classification gives rise to very imbalanced or underpopulated classes. \Cref{fig:methodology-1} shows the number of math problems per (primary) domain where we follow \cite{gao2024omni} in double- or triple-counting the multi-domain questions.  We made sure to delete the duplicate entries, e.g. some data entries had multiple domain trees but the same primary domain. Every domain-specific analysis in the paper follows this convention. Finally, we joined the Calculus and Pre Calculus class and deleted the Other class to obtain a more balanced domain distribution. Math problems are also classified according to difficulty level as presented in \Cref{fig:methodology-2}. We divide the data into difficulty tiers based on the quartiles of the difficulty distribution (without separating difficulty levels).

\textbf{OpenAI models}\quad We evaluate the performance of the several OpenAI models that are affordable for most users: gpt-4o-06-08-2024, o1-mini-12-09-2024, o3-mini-31-01-2025 medium (default) and o3-mini-31-01-2025 high. The o3-mini high model, instead of medium, is obtained by setting \texttt{reasoning\_effort} to high. We feed each model the math problems using the Batch API with the following vanilla prompt as user message: 
\begin{tcolorbox}[colframe=mygrey]
Solve the following problem. Enclose the final answer in a $\setminus \hspace{-1mm}\setminus\hspace{-1mm}\text{boxed}\{\{\}\}$ environment. Problem: \{problem\}
\end{tcolorbox}
Furthermore, we set \texttt{max\_completion\_tokens} limits of $25,000$ for o1-mini and o3-mini medium, and a $100,000$ token limit for o3-mini high. Each reasoning model refused to answer a few questions (flagged as invalid prompts), which were subsequently omitted from the analysis.

\textbf{Omni-Judge}\quad To correct the responses of the four OpenAI models on the Omni-MATH dataset, we employ another Large language model called Omni-Judge (\texttt{KbsdJames/Omni-Judge}). Omni-Judge is an efficient and low cost open-source math-evaluation model developed by the authors of \cite{gao2024omni}. The model is trained to assess the correctness of an answer generated by an LLM, given the problem and a reference answer (see \Cref{fig:sample-2}). Table $9$ in \cite{gao2024omni} shows that Omni-Judge is $91.78\%$ consistent with gpt-4o as a judge (who is $98\%$ consistent with human evaluators) and has almost a $100\%$ success rate of correctly parsing model generated answers. To judge the models' generated answers, we make requests to the chat completions endpoint of the \texttt{kbsdjames.omni-judge} API  by running the model in LM Studio. We use the same few-shot prompt as in \cite{gao2024omni} and set the \texttt{max\_new\_tokens} parameter to $300$. In the very few cases where Omni-Judge fails to parse the model output $(<1\%)$, we omit that question from the performance evaluation.

\begin{figure}[t]
    \centering
    \includegraphics{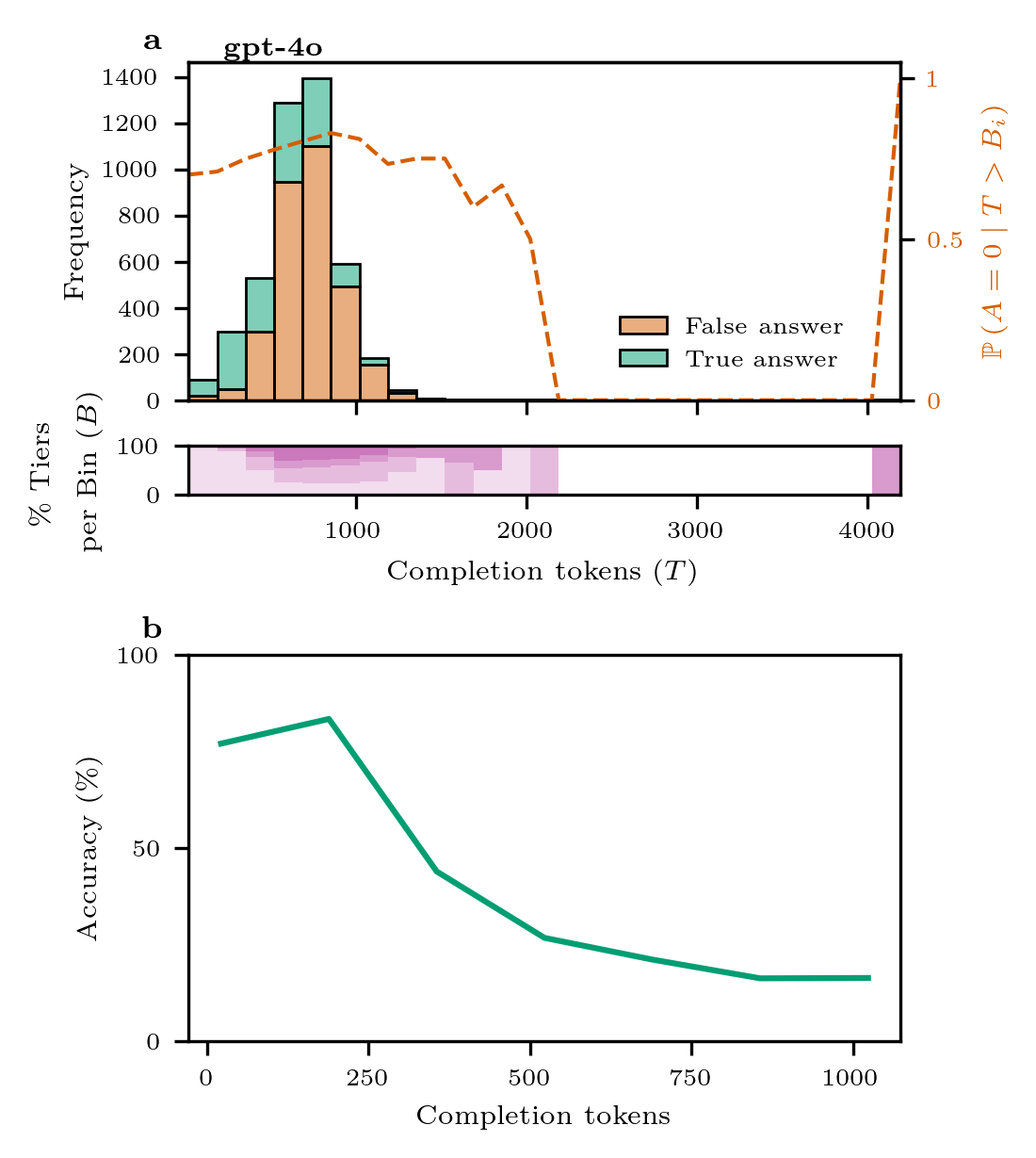}
    \caption{\textbf{Analysis of the completion token distribution, and the relationship between completion token usage and accuracy.} This figure shows that gpt-4o uses predominantly between $200$ and $1000$ completion tokens for answering the Omni-MATH problems. We also observe that shorter answers are more likely to lead to a correct final answer. Finally, the relative proportion of tier levels in each bin reveals a clear transition from a region where the majority of the questions come from the lowest tiers to a region where the majority of the questions come from the highest tiers (for bins with a sufficient amount of data points). \textbf{a,}
    The main panel of this plot displays a stacked histogram of the reasoning tokens used for correctly and incorrectly answered questions in the Omni-MATH dataset. The secondary $y$-axis depicts the probability that the model answers incorrectly given that the token count has surpassed the bin threshold (see \nameref{sec:Methods}). The subplot contains a filled histogram where the color opacity represents the difficulty level of the math questions (cfr. \Cref{fig:methodology-2}). \textbf{b,} Accuracy per reasoning token, computed by dividing the number of correctly answered questions by the total number of questions in each bin of the histogram depicted in \textbf{a}. Accuracy declines as completion token usage increases.
    }
    \label{fig:appendix-1}
\end{figure}

\textbf{Conditional probability}\quad The conditional probability appearing in \Cref{fig:main_3} and \Cref{fig:appendix-1} is computed using a full Bayesian model with uninformative priors (we assume that $\mathbb{P}(\text{False}) = \mathbb{P}(\text{True})=0.5$). In particular, we have that

\begin{align}
\begin{split}
        \mathbb{P}(\text{False} \mid T>B_i) &= \frac{\mathbb{P}(T>B_i \mid \text{False})}{\mathbb{P}(T>B_i \mid \text{False})+ \mathbb{P}(T>B_i \mid \text{True})}, 
\end{split}
\end{align}

where $\{T > B_i\}$ is the event that the number of tokens exceeds the right bin threshold and "False" indicates that the model answered incorrectly. Because $B_i$ can only take a finite number of values, we have that

\begin{align}
     \mathbb{P}(\text{False} \mid T>B_i) &= \sum_{k=i+1}^{n} \frac{\mathbb{P}(T \in B_k \mid \text{False})}{\mathbb{P}(T \in B_k \mid \text{False})+ \mathbb{P}(T \in B_k \mid \text{True})} = \sum_{k=i+1}^{n} \frac{|\text{False}\in B_k|}{|\text{False}\in B_k|+ |\text{True}\in B_k|},
\end{align}

which can be easily computed using the stacked histogram data. 

\clearpage
\textbf{Estimating effect sizes}\quad We use a logistic regression to estimate the effect of additional reasoning tokens on the probability of an accurate response on a question $Y_i$, while controlling for different levels of difficulty and domains. The regression takes the following form:
\begin{equation}\label{Regression equation}
\log \left( \frac{\Pr(Y_i = 1)}{\Pr(Y_i = 0)} \right) = \beta_0 + \beta_1 \text{tokens}_i + \underbrace{\sum_{k=1}^{K-1} \delta_k \; \text{difficulty tier}_{k(i)}}_{\substack{\text{difficulty fixed effects}}} + \underbrace{\sum_{m=1}^{M-1} \gamma_m \; \text{domain}_{m(i)}}_{\substack{\text{domain fixed effects}}},
\end{equation}
where $i$, $k$, and $m$ denote the question-response pair, the difficulty tier, and the domain, respectively. Moreover, $k(i)$ and $m(i)$ indicate that the difficulty tier $k$ and domain $m$ depend on the question-response pair $i$. The difficulty tier and domain fixed effects can be estimated by including dummy variables, which are equal to one if the difficulty tier or domain is equal to the difficulty tier or domain of the current question-response pair and equal to zero otherwise, with the exclusion of a reference category (i.e., $K-1$ and $M-1$). The reference category for difficulty tiers is the lowest difficulty Tier $1$ and for the domains it is Algebra. We obtain similar results when using the more fine-grained difficulty levels ($0-10$) instead of difficulty tiers.

To facilitate interpretation, we compute the Average Marginal Effect (AME) of additional reasoning tokens on the probability of an accurate response. Unlike the raw logistic regression coefficients, which are expressed in log-odds, the AME directly quantifies the effect of an additional token in probability terms. Specifically, it represents the average change in the probability of accuracy for a one-token increase, while holding difficulty tier and domain constant. By computing AMEs, we ensure that our estimates account for the full distribution of difficulty levels and domains, rather than relying on effects evaluated at a single reference point.

\clearpage
\section{Figures}

\vfill

\begin{figure}[h]
    \centering
    \begin{tcolorbox}[colframe=mygrey,fonttitle=\bfseries,title= Sample problem Omni-MATH]
\textbf{Domain: }Mathematics $\rightarrow$ Discrete Mathematics $\rightarrow$ Graph Theory

\textbf{Difficulty: }$9.0$

\textbf{Source: }China National Olympiad\newline

\textbf{Problem: }A table tennis club hosts a series of doubles matches following several rules:
\begin{enumerate}
  \item each player belongs to two pairs at most;
  \item every two distinct pairs play one game against each other at most;
  \item players in the same pair do not play against each other when they pair with others respectively.
\end{enumerate}

Every player plays a certain number of games in this series. All these distinct numbers make up a set called the \textit{set of games}. Consider a set $A=\{a_1,a_2,\ldots,a_k\}$ of positive integers such that every element in $A$ is divisible by $6$. Determine the minimum number of players needed to participate in this series so that a schedule for which the corresponding \textit{set of games} is equal to set $A$ exists. \newline

\textbf{Answer: }$\frac{1}{2}\max A + 3.$\newline

\textbf{Solution: } To determine the minimum number of players needed to participate in the series such that the set of games is equal to the set \(A\), we start by analyzing the problem through graph theory.

Consider a graph \(\mathcal{G}\) where each vertex represents a player and an edge between two vertices represents a pair of players. According to the problem's conditions:
\begin{enumerate}[label=(\roman*)]
  \item Each player belongs to at most two pairs.
  \item Every two distinct pairs play one game against each other at most.
  \item Players in the same pair do not play against each other when they pair with others respectively.
\end{enumerate}

Given these conditions, each vertex in \(\mathcal{G}\) can have a degree of at most 2. This implies that \(\mathcal{G}\) can be decomposed into disjoint cycles, paths, and isolated vertices. Let \(\max A = a\). We need to ensure that there is a player who plays \(a\) games, and each element in \(A\) is divisible by 6. To achieve this, we need to construct a graph where the number of games each player plays corresponds to the elements in \(A\). \#\#\# Proof of optimality: Let \(x\) be a vertex with the maximum number of games \(a\). This vertex \(x\) must be connected to at least one other vertex. If \(x\) has only one edge, it must have \(a\) games on it. If \(x\) has two edges, each edge must have at least \(\frac{a}{2}\) games. Let \(xy\) be an edge with \(\geq \frac{a}{2}\) games. The remaining vertices must account for at least \(\frac{a}{2}\) edges, ensuring that there are at least \(\frac{a}{2} + 2\) vertices. However, due to the constraints, there must be at least \(\frac{a}{2} + 3\) vertices to satisfy all conditions.
\#\#\# Construction: To construct such a graph, we can use clusters of three players, each forming pairs within the cluster. This ensures that each player in a cluster plays 6 games per edge. By assigning each cluster to a vertex in a graph \(\mathcal{G}\) with \(b\) vertices, where $b = \frac{1}{6}\max A + 1$,
we can ensure that the set of games corresponds to \(A\). Thus, the minimum number of players needed is: $\frac{1}{2} \max A + 3 $. The answer is: $$\boxed{\frac{1}{2} \max A + 3}$$
\end{tcolorbox}
    \caption{\textbf{Sample problem from the Omni-MATH dataset.} The Omni-MATH dataset consists of $4428$ Olympiad-level math problems together with an exact answer, a written out solution and metadata Domain, Difficulty and Source.}
    \label{fig:sample-1}
\end{figure}

\vfill

\clearpage
\begin{figure}[t]
    \centering
    \scalebox{1}{
    \begin{tcolorbox}[colframe=mygrey,fonttitle=\bfseries,title= Sample output Omni-judge]
\textbf{Problem: }A table tennis club hosts a series of doubles matches following several rules:
\begin{enumerate}[label=(\roman*)]
  \item each player belongs to two pairs at most;
  \item every two distinct pairs play one game against each other at most;
  \item players in the same pair do not play against each other when they pair with others respectively.
\end{enumerate}

Every player plays a certain number of games in this series. All these distinct numbers make up a set called the \textit{set of games}. Consider a set $A=\{a_1,a_2,\ldots,a_k\}$ of positive integers such that every element in $A$ is divisible by $6$. Determine the minimum number of players needed to participate in this series so that a schedule for which the corresponding \textit{set of games} is equal to set $A$ exists. \newline

\textbf{Answer: }$\frac{1}{2}\max A + 3.$\newline

\textbf{Model generation: }We will show that one may `hit' a prescribed set $A = \{a_1, a_2, \ldots, a_k\}
$ (where each $a_i$ is a positive multiple of $6$) by the following construction, and that no schedule may use fewer than $$\frac{a_1+a_2+\cdots+a_k}{3} + 2k \quad (1)$$ players. [...]
In any such schedule every pair (team) plays some number of games. (Rule (ii) means that once we have fixed a pair $P$ the `opponent--teams' of $P$ form a set; we denote by $d(P)$ the number of games in which $P$ plays.) Then a given player’s `score' (the number of games in which he participates) is determined by whether he belongs to one pair (in which case his score is the degree of that pair) or to two pairs (in which case his score is the sum of the two degrees). (Note that by rule (iii) if a player belongs to two pairs then those two pairs never meet so that the numbers add!) Thus if we `read off' the numbers of games played by the players (omitting any repetitions) we obtain a set - called the `set of games'. In our problem we wish this set to be exactly the given $A$.

An important idea is that players who belong to two pairs are `economical' in that one player now serves to `cover' two teams. (In a `non-economical' schedule each pair would contribute two `single--members'.) One may show that in a schedule in which as many players as possible are `doubles' (that is, belong to two teams) one may `control' the scores so that in fact every player’s number is of the form $d(P) \text{ or } d(P)+d(Q)$ (with $d(P)$ and $d(Q)$ coming from a suitable graph on the pairs). One may then prove the following two claims:\newline

\textbf{Claim 1.} If the schedule uses only double--players then the total number $n$ of players is exactly the number $T$ of pairs, and a routine ``pairing'' (that is, a perfect matching) of the teams shows that one must have $n \ge 2k$ (since one obtains exactly $n/2$ sums, one for each game--of--a--double, and these must contain the $k$ given numbers.) \newline

\textbf{Claim 2.} Since a team’s degree is at most $(n-1)$ the maximum score obtainable by a double is at most $2(n-1)$. In order to be able to ``reach'' a number $a_i$ we must have $2(n-1) \ge a_i \quad \text{for every } i$,
so that in particular $n \ge \frac{1}{2}\max\{a_i\} + 1$. A careful double--counting (namely, counting the total number of games by summing the scores of players in two different ways) shows that one must have $n\ge \frac{a_1+a_2+\cdots+a_k}{3} + 2k$.
Moreover one may show by an inductive construction that equality is attainable. One acceptable answer to the problem is therefore to prove that the minimum number of players needed is $\frac{a_1+a_2+\cdots+a_k}{3} + 2k$. An answer equivalent to (1) is completely acceptable. Thus, our final answer is
\[
\boxed{\frac{a_1+a_2+\cdots+a_k}{3}+2k}.
\] \newline

\textbf{Omni-Judge: }\#\# Student Final Answer: $\frac{a_1+a_2+\cdots+a_k}{3}+2k$. \#\# Equivalence Judgement: FALSE \#\# Justification: The student's answer is $\frac{a_1+a_2+\cdots+a_k}{3}+2k$,
which does not match the reference answer of $\frac{1}{2}\max A + 3$. The discrepancy lies in the treatment of the maximum value of $A$ and the constant term. In the student's solution, they consider the average of all elements in set $A$, whereas the reference answer focuses on half of the maximum element in $A$ plus an additional constant term. This difference indicates that the student's answer does not capture the same meaning as the reference answer. === report over ===

\end{tcolorbox}}
    \caption{\textbf{Sample of the Omni-Judge output when presented with a problem, a reference answer and an o3-mini (h) generated solution.} The Omni-Judge model is prompted as in \cite{gao2024omni}, which is to repeat the `student's' final answer, asses the equivalence of this answer with the reference answer and provide complementary justification. We omit the part where o3-mini (h) repeats the problem specifications, indicated by [...].}
    \label{fig:sample-2}
\end{figure}

\clearpage

\clearpage

\clearpage

\clearpage
\begin{figure}[t]
    \centering
    \includegraphics{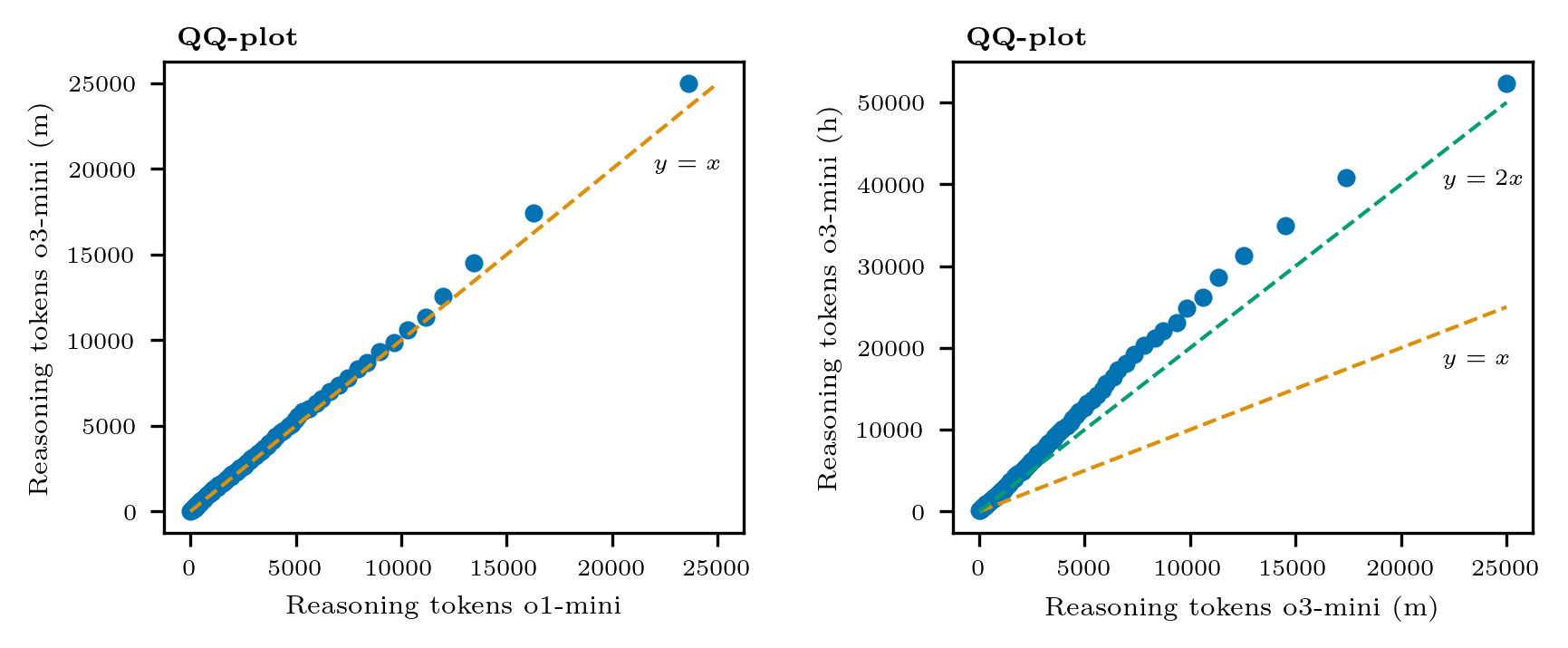}
    \caption{\textbf{Comparison of the distribution of reasoning tokens between o1-mini and o3-mini (m), and between o3-mini (m) and o3-mini (h) for \emph{correctly} answered problems in the Omni-MATH dataset.} This figure compares the token distribution of three OpenAI reasoning models by means of a QQ-plot. We observe that o1-mini and o3-mini (m) have an almost identical reasoning token distribution when we only consider the correctly answered questions. The token distribution of o3-mini (h) is a linearly scaled version of the distribution of o3-mini (m) with a factor slightly larger than $2$. }
    \label{fig:appendix-3}
\end{figure}

\clearpage
\begin{figure}[t]
    \centering
    \includegraphics{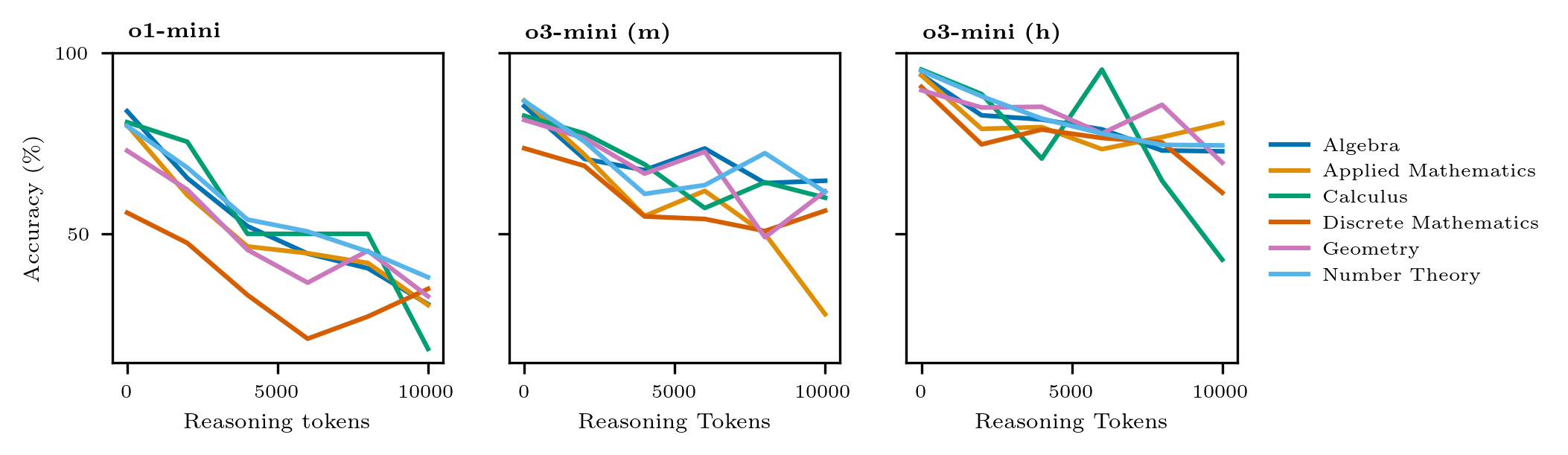}
    \caption{\textbf{Stratification of \Cref{fig:main_4}a for the mathematical domains of the Omni-MATH dataset.} This figure displays a stratification of \Cref{fig:main_4}a by mathematical domain.
    In general, accuracy decreases within the domains as the use of reasoning tokens increases. }
    \label{fig:appendix-2}
\end{figure}

\clearpage
\section{Tables}

\vfill
\begin{figure}[h]
    \centering
\centering
\renewcommand{\arraystretch}{1.2}
\footnotesize 
\begin{tabular}{{@{\extracolsep{4pt}}l L{2cm} L{2cm} L{2cm}@{}}}
\toprule
 & o1-mini & o3-mini (m) & o3-mini (h) \\
\midrule
\midrule
\textbf{without controls} \\
Tokens & $-1.85\mathrm{e}{-4}^{***}$ & $-1.25\mathrm{e}{-4}^{***}$ & $-5.77\mathrm{e}{-5}^{***}$ \\\\
Constant & $1.19^{***}$ & $1.53^{***}$ & $1.87^{***}$ \\\\

\textbf{with controls} \\
Tokens & $-1.61\mathrm{e}{-4}^{***}$ & $-1.08\mathrm{e}{-4}^{***}$ & $-5.10\mathrm{e}{-5}^{***}$ \\\\

Difficulty tier 2 & $-0.53^{***}$ & $-0.36^{***}$ & $-0.20^{**}$ \\
Difficulty tier 3 & $-0.74^{***}$ & $-0.56^{***}$ & $-0.37^{***}$ \\
Difficulty tier 4 & $-1.08^{***}$ & $-0.70^{***}$ & $-0.63^{***}$ \\\\

Applied Mathematics & $-0.41^{***}$ & $-0.34^{***}$ & $-0.37^{***}$ \\
Calculus & $0.13$ & $0.03$ & $0.09$ \\
Discrete Mathematics & $-0.86^{***}$ & $-0.50^{***}$ & $-0.41^{***}$ \\
Geometry & $-0.46^{***}$ & $-0.21^{**}$ & $-0.26^{**}$ \\
Number Theory & $0.02$ & $0.04$ & $0.08$ \\
Other & $0.47$ & $0.02$ & $-0.03$ \\\\
Constant & $1.93^{***}$ & $2.02^{***}$ & $2.25^{***}$ \\
\midrule
N & $5,535$ & $5,531$ & $5,526$ \\\\
McFadden's pseudo-$R^2$ & $0.11$ & $0.06$ & $0.06$ \\
(without controls) \\\\
McFadden's pseudo-$R^2$ & $0.15$ & $0.07$ & $0.08$ \\
\bottomrule
\end{tabular}
\label{tab:regression}

   \caption{\textbf{Logistic regression models to estimate the effect size of the number of tokens on response accuracy.} We use a logistic regression to estimate the effect of additional reasoning tokens on the probability of an accurate response on a question, while controlling for different levels of difficulty and domains. Estimates are from a logistic regression (Eq. \ref{Regression equation}) fit by maximum likelihood, with robust (Huber–White) standard errors to account for potential heteroskedasticity. The significance levels are for a two-sided Wald test with a null hypothesis of the regression coefficient being equal to zero (\textsuperscript{***}$p<0.01$, \textsuperscript{**}$p<0.05$, \textsuperscript{*}$p<0.1$).}
    \label{fig:enter-label}
\end{figure}

\vfill

\end{document}